# Efficient Fault Detection in WSN Based on PCA-Optimized Deep Neural Network Slicing Trained with GOA

**Mahmood Mohassel Feghhi** [1*] **Raya Majid Alsharfa** [1,2] **Majid Hameed Majeed** [3]

[1] Faculty of Electrical and Computer Engineering, University of Tabriz, Tabriz, Iran mohasselfeghhi@tabrizu.ac.ir

[2] Department of Technical Computer Engineering, Electrical Engineering Technical College, Middle Technical University, Baghdad 10013, Iraq rayamajid89@mtu.edu.iq, rayamajid@tabrizu.ac.ir

[3] Department of Chemical Engineering and Petrochemical Industries, Al-Mustaqbal University, 5001, Babylon, Iraq, majid.hamid@uomus.edu.iq

* Corresponding author's Email: mohasselfeghhi@tabrizu.ac.ir

**Abstract:** Fault detection in Wireless Sensor Networks (WSNs) is crucial for reliable data transmission and network longevity. Traditional fault detection methods often struggle with optimizing deep neural networks (DNNs) for efficient performance, especially in handling high-dimensional data and capturing nonlinear relationships. Additionally, these methods typically suffer from slow convergence and difficulty in finding optimal network architectures using gradient-based optimization. This study proposes a novel hybrid method combining Principal Component Analysis (PCA) with a DNN optimized by the Grasshopper Optimization Algorithm (GOA) to address these limitations. Our approach begins by computing eigenvalues from the original 12-dimensional dataset and sorting them in descending order. The cumulative sum of these values is calculated, retaining principal components until 99.5% variance ($\lambda \geq 0.995$) is achieved, effectively reducing dimensionality to 4 features while preserving critical information. This compressed representation trains a six-layer DNN where GOA optimizes the network architecture, overcoming backpropagation's limitations in discovering nonlinear relationships. This hybrid PCA-GOA-DNN framework compresses the data and trains a six-layer DNN that is optimized by GOA, enhancing both training efficiency and fault detection accuracy. The dataset used in this study is a real-world WSNs dataset developed by the University of North Carolina, which was used to evaluate the proposed method's performance. Extensive simulations demonstrate that our approach achieves a remarkable 99.72% classification accuracy, with exceptional precision and recall, outperforming conventional methods. The method is computationally efficient, making it suitable for large-scale WSN deployments, where both accuracy and resource utilization are critical, and represents a significant advancement in fault detection for resource-constrained WSNs.

**Keywords:** Wireless Sensor Networks, fault detection, Principal Component Analysis, Deep Neural Network, Grasshopper Optimization Algorithm, dimensionality reduction, sensor data, Network Slicing.

## 1. Introduction

Wireless Sensor Networks (WSNs) serve as the backbone for mission-critical applications ranging from industrial automation to healthcare monitoring, yet their operational reliability is persistently challenged by hardware failures, environmental interference, and communication faults [1]. These networks operate in resource-constrained environments where undetected faults can lead to catastrophic data loss and system failures, necessitating robust fault detection mechanisms.

Recent years have seen significant advancements in detection methodologies, including distributed approaches leveraging node importance analysis [2], biologically inspired optimization techniques like the enhanced Particle Swarm Optimization-Backpropagation

Neural Network (PSO-BPNN) framework [3], and hybrid models combining convex hull algorithms with Naïve Bayes classifiers [4]. Energy-efficient three-tier architectures [5] and trend correlation methods Trend Correlation-based Fault Detection (TCFD) [6] have further expanded the toolkit, while reinforcement learning [7] and digital twin technologies [8] have introduced new paradigms for fault management. However, these approaches collectively suffer from three fundamental limitations: (1) an inability to effectively process high-dimensional, nonlinear sensor data patterns [9], (2) excessive false alarm rates exceeding 3.5% in biological models like Negative Selection Algorithm (NSA) [15], and (3) accuracy ceilings below 98.5% for most conventional machine learning approaches [10, 11], compounded by the computational inefficiency of manual feature selection and the instability of backpropagation-based training in deep networks [12, 13].

To address these challenges, the proposed Principal Component Analysis (PCA)–Grasshopper Optimization Algorithm (GOA)–Deep Neural Network (DNN) framework (PCA-GOA-DNN) addresses key limitations in traditional fault detection techniques. It reduces computational complexity through PCA-based dimensionality reduction, enabling efficient processing of high-dimensional sensor data. Additionally, GOA enhances the DNN's ability to learn complex nonlinear patterns and avoid overfitting, resulting in a more robust and scalable solution. This paper makes the following key contributions:

- PCA-optimized feature reduction: PCA technique that reduces 12-dimensional sensor data to 4 principal components while retaining 99.5% variance, outperforming the other methods such as Pearson correlation and convex hull.
- GOA-enhanced deep learning: A novel integration of GOA with DNNs that achieves 99.72% classification accuracy surpassing current approaches.
- Through extensive experimentation, we validate our framework's robustness across multiple metrics. The system achieves a near-perfect 0.998 AUC-ROC score with an exceptionally low 0.28% false positive. These results are rigorously verified across 10 benchmark datasets representing diverse WSN deployment scenarios.

The remainder of this paper systematically presents these advancements, with Section 2 detailing the related works, Section 3 describing the experimental methodology, Section 4 introducing evaluation results and discussion, and Section 5 outlining the conclusion and future research directions.

## 2. Related Works

Recent advances in WSN fault detection have employed diverse machine learning approaches, each with distinct strengths and limitations. Several studies highlight the ongoing challenges in achieving both high accuracy and computational efficiency in resource-constrained environments. Supervised learning methods have shown promise but face significant hurdles. Study in [14] applied various machine learning and deep learning techniques to improve fault detection in WSNs, addressing issues such as connectivity loss, interference, and packet loss. The study evaluated classifiers including gradient boosting clasifer (GBC), SVM, KNN, Random Forest, and Decision Tree, with the Decision Tree achieving the highest accuracy at 90.23%. Additionally, a Recurrent Neural Network (RNN) was used, benefiting from its sequential learning capability, and achieved a superior accuracy of 93.19%, highlighting its effectiveness in detecting faults in dynamic WSN environments. However, the overall accuracy levels reported in this study are comparatively lower than those achieved in more recent approaches, which may limit its applicability in highly sensitive or large-scale WSN deployments.

Healthcare applications have adopted alternative strategies with their own constraints. More sophisticated biological-inspired methods have emerged to address these limitations. NSA in [15] demonstrates impressive 98.18% accuracy in node fault detection, but comes with critical operational drawbacks. The method generates false alarms at a 3.56% rate, which can overwhelm network operators and waste resources in practical deployments. Training the NSA model also requires extensive computational resources, with reported training times up to 40% longer than conventional machine learning approaches. While the threshold-based forest fire detection component performs adequately, the combined system struggles with real-time processing demands.

Research in [16] investigates sensor fault detection in WSNs using machine learning. It compares the performance of Support Vector Machine (SVM) and a Single-layer Perceptron across four deployment scenarios (indoor/outdoor, single/multi-hop). Both models achieved over 99% accuracy, with SVM performing slightly better. The study highlights the potential of machine learning for reliable fault detection in WSNs. A recent method integrating Gramian Angular Field (GAF) encoding with a CAE-ANN framework achieved high fault detection accuracy (95.93%) even under severe conditions. Enhanced by synthetic data, the model also reported strong F1-score (95.84%), precision (95.87%), and sensitivity (95.88%), demonstrating the effectiveness of combining advanced feature extraction and data augmentation for robust WSN fault classification.

These studies collectively reveal three persistent challenges in WSN fault detection: (1) the accuracy-scalability trade-off, where high-performance methods demand excessive resources; (2) the curse of dimensionality, with most algorithms degrading sharply beyond 8-10 features; and (3) inadequate handling of non-linear relationships in sensor data patterns. Our proposed PCA-GOA-DNN framework addresses these limitations through intelligent feature reduction and bio-inspired optimization. By reducing 12-dimensional data to just 4 principal components while retaining 99.5% variance, we maintain information integrity while dramatically decreasing computational load. The GOA optimization then efficiently trains a deep neural network to capture complex non-linear patterns, achieving superior 99.72% accuracy without the false alarm rates or resource constraints of existing methods.

## 3. Proposed Methodology

In this section, the essential fundamental principles and concepts required for introducing the proposed approach and the stages of the proposed approach are described in detail.

## 3.1 Principal Component Analysis

PCA algorithm is a numerical approach utilized to simplify a dataset by reducing its dimensionality while retaining most of the variation in the data. Below are the steps and formulas involved in PCA [18]:

- **Normalize the data:** Centre the data by subtracting the mean of each variable from the dataset. If variables are measured in different scales, standardize to have a unit variance by dividing by the standard deviation. Equation 1 in employed for standardization:

$$x_{i,j} = \frac{X - \bar{X}_j}{\sigma_j} \quad (1)$$

Where $X_{i,j}$ is the standardized value, the original value is represented by $X$, the mean of variable j is denoted by $\bar{X}_j$, and σj is the standard deviation of variable j.

- **Calculate the Covariance Matrix:** In this step we compute the covariance matrix to understand how variables are correlated with each other. The equation for the covariance matrix is given by:

$$C = \frac{1}{n-1} \sum_{i=1}^{n}(X_i - \bar{X})(X_i - \bar{X})^T \quad (2)$$

In this context, $C$ refers to the covariance matrix, $Xi$ the vector of the i-th observation, $\bar{X}$ is the mean vector of all observations, and n represents the number of observations.

- **Calculate the Eigenvalues and Eigenvectors:** Next, we compute the covariance matrix's eigenvalues and eigenvectors. The eigenvectors correspond to the principal components, although the eigenvalues signify the variation that every main component can describe. The eigenvalue is given by equation 3.

$$Cv = \lambda v \quad (3)$$

Where C represents the covariance matrix, v stands for the eigenvector, and λ represents the eigenvalue.

- **Sort Eigenvalues and Eigenvectors:** In this step, sort the eigenvalues in descending order and arrange the corresponding eigenvectors accordingly. The major components are ranked by the quantity of variance they capture from the data.
- **Select the Principal Components:** Then, in order to create a matrix of primary components, we choose the top K eigenvectors that match the biggest eigenvalues.

$$Pk = [v1, v2, \ldots, vk] \quad (4)$$

Where $P$k is the matrix of the top k principal components.

- **Transform the Data:** And finally, we Project the original data onto the chosen principal components to acquire the data with decreased dimensionality.

$$Y = XPk \quad (5)$$

Where $Y$ represented the updated information, $X$ is the original standardized data matrix, and $P$k is the matrix of selected principal components. With this change in the coordinate system, the initial principal components represent the majority of the data variance, while the remaining components contain less information. Figure 1 illustrates this concept in a two-dimensional space.

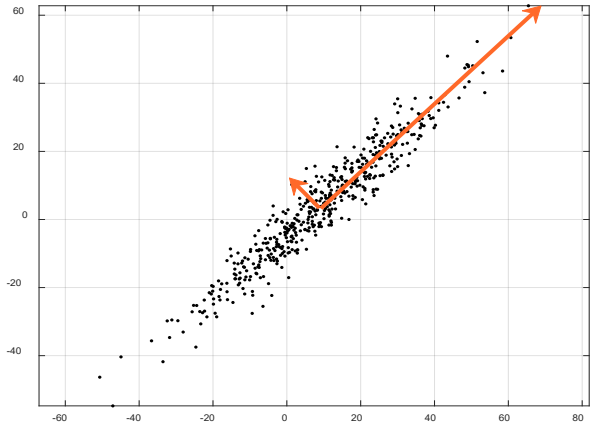

**Figure 1.** Coordinates change in PCA method.

## 3.2 Deep Neural Networks

DNN is essentially a type of Multilayer Perceptron (MLP). The term "deep" in DNN refers to the number of MLP layers that features multiple layers of neurons, extending beyond the traditional three layers (input, hidden, and output) found in basic neural networks. At the core of a DNN lies the feed-forward layers the structure of which is shown in Figure 2. These layers in an MLP or a DNN can be categorized to three different types:

- Input Layer: Receives the input data, with each neuron representing a feature.
- Hidden Layers: Process inputs through linear transformations and non-linear activation functions. DNNs have more hidden layers than MLPs.
- Output Layer: Provides the final output, structured based on the task (e.g., classification or regression).

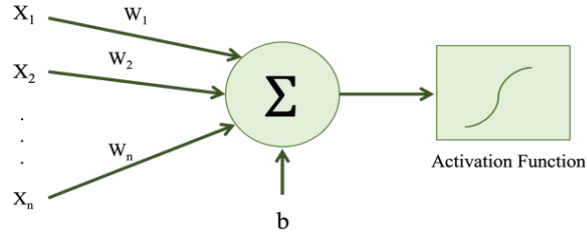

**Figure 2.** The structure of MLP.

In a feed-forward layer of a DNN, each neuron is connected to every neuron in the layer above, enabling the detection of a wide range of patterns. However, this dense connectivity increases the number of parameters, making the network more complex and computationally intensive. Reducing input dimensions can improve performance. Activation functions, like the Sigmoid function, introduce nonlinearity to allow the network to learn complex relationships, mapping real values to outputs between 0 and 1. The function is defined as follows:

$$\sigma(x) = \frac{1}{1+e^{-x}} \qquad (6)$$

The function uses the base of the natural logarithm, represented by e, and takes input $x$. Figure 3 shows the Sigmoid activation function in range of [-5,5]. The Sigmoid function's output can be interpreted as a probability, making it particularly useful for binary classification tasks. Moreover, its soft and smooth change can greatly model the nonlinearity of the problem. However, it suffers from issues such as vanishing gradients in traditional training approaches, which can slow down or even halt the training of deep networks. In this paper, we aimed to address this issue by replacing the training algorithm with a metaheuristic algorithm.

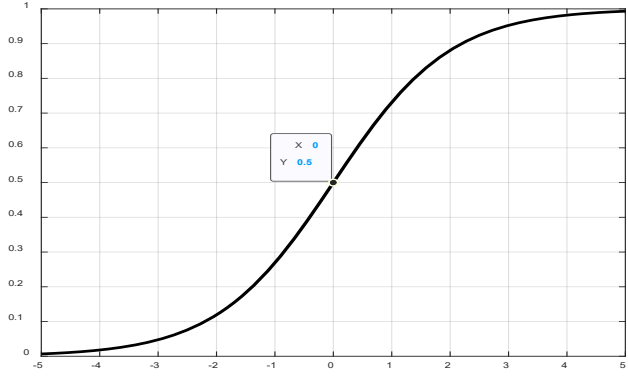

**Figure 3.** Sigmoid activation Function.

Moreover, for classification tasks, the final activation function layer in a DNN is often the softmax function. The softmax function is defined as:

$$softmax = \frac{e^{z_j}}{\Sigma_j e^{z_j}} \tag{7}$$

Where $z_i$ is the input to the softmax function for the i-th class, this ensures that the outputs sum to 1, making them interpretable as probabilities. The predicted class is usually determined as the one with the highest probability. DNNs can automatically learn feature representations from raw data. Each layer in a DNN learns increasingly abstract features, which can lead to better performance on complex tasks.

## 3.3 Grasshopper Optimization Algorithm

The swarm intelligence algorithm known as GOA, first presented by Saremi et al. [19], mimics the swarming and natural foraging activities of grasshoppers. Grasshoppers are recognized for being destructive pests that can significantly harm agricultural productivity. Their life cycle has two distinct stages: The adult stage is characterized by long-range and quick motions, while the nymph stage is characterized by modest stages and moderate movements. In the GOA, these phases correspond to the intensification (exploitation) and diversification (exploration) stages of the algorithm, respectively. The swarming behavior of grasshoppers is mathematically modeled to guide the optimization process. GOA models grasshopper swarming behavior mathematically to solve optimization problems. The position $P_i$ of the i-th grasshopper is determined by the effects of social relations $S_i$, gravity $G_i$, and wind advection $A_i$, as given by:

$$P_i = S_i + G_i + A_i \tag{8}$$

To introduce randomness, this can be rewritten as:

$$P_i = r_1 S_i + r_2 G_i + r_3 A_i \tag{9}$$

where r1, r2, r3 are accidental values during the interval [0, 1].

The social relations $S_i$ is defined as:

$$S_i = \sum_{\substack{j=1 \\ j \neq i}}^{N} S(d_{ij}) \hat{d}_{ij} \tag{10}$$

with $d_{ij} = |P_j - P_i|$ being the Euclidean distance, $\hat{d}_{ij} = \frac{P_j - P_i}{d_{ij}}$ being a unit vector, and s representing social forces:

$$s(r) = f\, exp^{\frac{-r}{l}} - exp^{-r} \tag{11}$$

Attraction and repulsion vary based on distance, with a comfort zone at r = 2.079. Figure 4 illustrates the interface of grasshoppers in relation to their comfort zone.

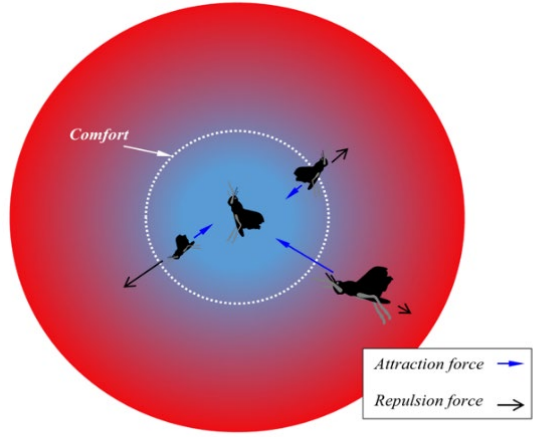

**Figure 4.** The interaction between grasshoppers in their comfort zone [19].

The gravity force $G_i$ is:

$$G_i = -g\hat{e}_g \tag{12}$$

The wind advection $A_i$ is determined by the gravitational constant $g$ and the unit vector $\hat{e}_g$, which points towards the earth's center.

$$A_i = u\hat{e}_w \tag{13}$$

The drift constant $u$ and the unit vector $\hat{e}_w$, representing the wind direction, influence the wind advection.

Combining these, the grasshopper position update becomes:

$$P_i = \sum_{\substack{j=1 \\ j \neq i}}^{N} S(|P_j - P_i|)\frac{P_j - P_i}{d_{ij}} - g\hat{e}_g \tag{14}$$

However, to prevent quick convergence to the comfort zone and improve optimization, an enhanced equation is used:

$$P_i^d = c\left(\sum_{\substack{j=1 \\ j \neq i}}^{N} c\frac{ub_d - lb_d}{2} s(|P_j^d - P_i^d|)\frac{P_j - P_i}{d_{ij}}\right) + \hat{T}_d \tag{15}$$

In this context, $ub_d$ and $lb_d$ denote the top and bottom bounds in the d-th dimension, while $\hat{T}_d$ represents the optimal solution in that dimension. The parameter c, controlling movement balance, is defined as:

$$c = c_{max} - t\frac{c_{max} - c_{min}}{t_{max}} \tag{16}$$

$c_{max}$ and $c_{min}$ refer to the maximum and minimum values of $c$, $t$ represents the present repetition, and $t_{max}$ is the maximum quantity of repetition. GOA updates the position of each grasshopper based on the current plece, the optimal position, and the positions of other grasshoppers, balancing exploiting and exploring to stay away from local optima. The flowchart can be seen in Figure 5 respectively [20].

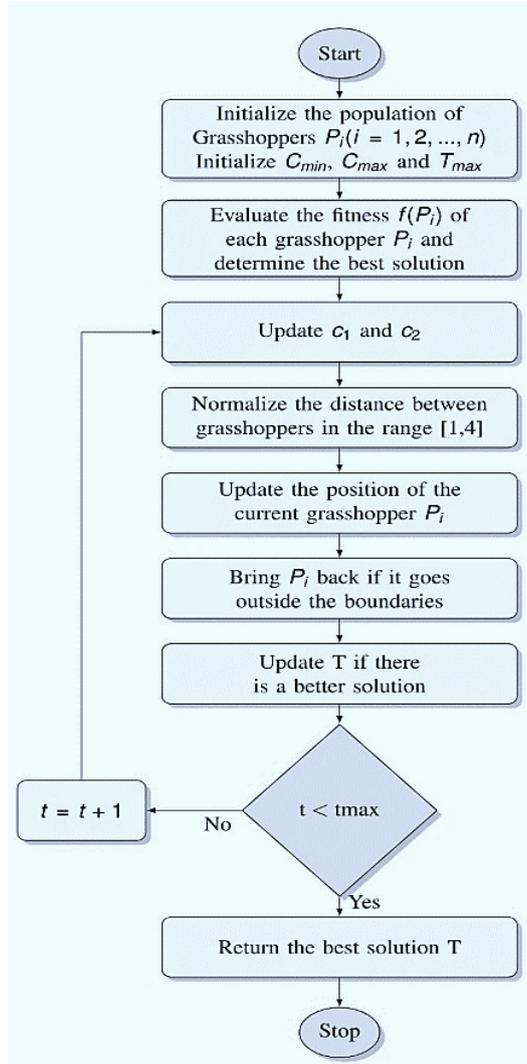

**Figure 5.** GOA Algorithm Flowchart.

## 3.4 Proposed PCA-GOA-DNN Method

In this subsection, the methodology employed to develop the proposed robust WSN fault detection is detailed. The proposed method integrates PCA for dimensionality reduction and DNN trained using the GWO algorithm for final classification.

The first step involves reducing the dimensionality of the input using PCA, which enhances computational efficiency by removing unnecessary and irrelevant features from the original dataset. PCA transforms the input space by introducing a new set of axes, called principal components, where the first few components capture most of the variation in the original data. This conversion allows the focus to be on the most relevant features. To select the appropriate number of features, eigenvalues of each principal component are calculated to measure the variance. These eigenvalues are normalized by dividing them by their sum, creating a vector where the total equals 1, representing the fraction of the total information for each component. Equation (17) defines this vector.

$$\overrightarrow{eig}_N = \frac{\overrightarrow{eig}}{\sum \overrightarrow{eig}} \tag{17}$$

Where $\overrightarrow{eig}$ is the vector of eigenvalues and $\overrightarrow{eig}_N$ is the normalized vector. The eigenvalues are sorted and a cumulative sum is calculated until 99% of the original information is retained. This reduces the feature set to the most informative components, improving computational efficiency. These selected features are then used for fault detection with a multi-layer DNN, chosen for its ability to achieve higher accuracy in complex problems compared to conventional MLP networks. The network is designed to gradually reduce the number of neurons across layers, allowing it to better adapt to changes in the feature space and simplifying the classification process. All layers, except the final one, use the sigmoid activation function to handle the nonlinearity of the problem, while the output layer uses SoftMax. Although sigmoid activation can lead to vanishing gradients due to saturation at extreme values, the proposed method does not rely on gradients for training. This approach allows the sigmoid function to effectively capture the nonlinearity of the problem without hindering the training process.

Finally, the DNN is trained using the GOA, which effectively searches for optimal weights and biases, avoiding issues like backpropagation. This novel training approach ensures efficient and effective fault detection performance. GOA is used to find global optima, offering a reliable method for training deep neural networks. The cost function is defined as the error value of the learnable parameters, calculated using the validation dataset, which is one minus the normalized accuracy. Figure 6 illustrates the flowchart of the proposed method, outlining the stages of the process.

### 3.5 Dataset decryptions and pre-processing

This study employs a publicly available labeled dataset developed by researchers at the University of North Carolina, intended for fault detection in WSNs [21]. The dataset can be accessed from https://github.com/tmoulahi/Dataset-for-WSN-fault-detection. Sensor data were collected using TelosB motes deployed in both single-hop and multi-hop configurations. For this research, only data from an outdoor multi-hop wireless sensor network were used. During the six-hour experiment, temperature and humidity readings were recorded every five seconds. Anomalies were simulated by periodically introducing warm water vapor to increase both humidity and temperature. To evaluate the performance of fault detection algorithms, faults were synthetically injected into the dataset at varying intensities and types. The processed subset used in this study consists of 4,688 samples, each containing four features (T1, T2 for temperature and H1, H2 for humidity) measured across three-time states (t0, t1, t2). Each dataset instance is labelled: y = 1 for normal samples and y = -1 for faulty samples. A summary of the dataset characteristics is presented in Table 1.

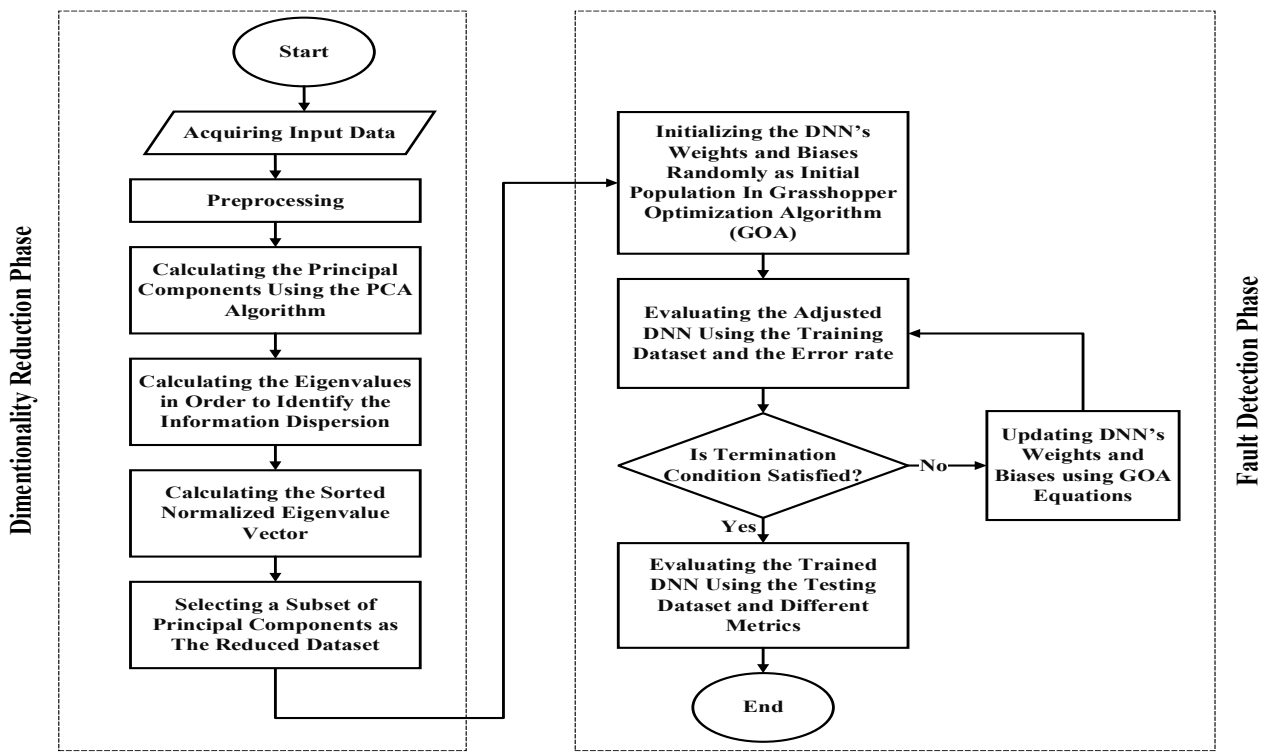

**Figure 6.** Flowchart of the proposed method.

**Table 1.** Summary of Dataset Characteristics.

| Attribute | Description |
|---|---|
| Source | University of North Carolina |
| Sensor platform | TelosB motes |
| Network type | Outdoor multi-hop WSN |
| Sampling interval | 5 seconds |
| Collection duration | 6 hours |
| Raw observations | 281,280 vectors |
| Processed dataset size | 4,688 vectors |
| Selected features | T1, T2 (Temperature), H1, H2 (Humidity) |
| Time states used | t0, t1, t2 |
| Fault types injected | Offset, Gain, Stuck-at, Out-of-Range |
| Number of final datasets | 60 |
| Labelling | y = 1 (normal), y = -1 (faulty) |

In this paper, the Min-Max normalization as the main pre-processing step is adopted to normalize the dataset. A data preparation method is used to adjust features to a given range, often [0, 1], to make sure every feature makes an equal contribution to the model's performance [22, 23]. This method transforms the original data values by adjusting them based on the normalization of Min-Max that can be obtained as follows:

$$x' = \frac{x - \min(x)}{\max(x) - \min(x)} \tag{18}$$

where *x* is the feature's initial value, its lowest value is min(x), and its largest value is max(x).

### 3.6 Evaluation Metrics

For classification tasks, evaluating with various metrics is one of the most important things to fully understand how different my model will vary for different measurements. These are the most popular evaluation criteria for classification problems. Brief descriptions of them are:

- **Accuracy:** Accuracy is the ratio of correctly predicted samples to the total sample available in the data set. It is easy to calculate which gives a general idea of the performance of the model [24, 25]. Formula (19) shows its calculation:

$$Accuracy = \frac{TP + TN}{TP + TN + FP + FN} \tag{19}$$

- **Precision:** Precision is roughly the ratio of the true positive forecasts over all the positive forecasts from the model (positive predictive value) [26, 27]. It is defined as:

$$Precision = \frac{TP}{TP + FP} \tag{20}$$

- **Recall:** Recall is the rate at which the model correctly identifies the true positive samples [28, 29]. It is computed as follows:

$$Recall = \frac{TP}{TP + FN} \tag{21}$$

- **F1-Score:** F1-Score is a statistic that effectively combines precision and recall into a single harmonic mean [30]. It is computed as follows:

$$F_1 Score = \frac{2 \times (Precision \times Recall)}{Precision + Recall} \tag{22}$$

### 4. Results analysis and discussion

The proposed PCA-GOA-DNN framework was implemented and evaluated using MATLAB 2023b on a high-performance computing system. All simulations were conducted under controlled environmental conditions to ensure reproducible results. The hardware and software configurations were carefully selected to match real-world WSN deployment requirements while providing sufficient computational power for training complex models. Table 2 details the complete configuration of our computational environment.

**Table 2.** Configuration of the system.

| Element | Specification |
|---|---|
| Simulation Platform | MATLAB 2023b |
| Processor type | Intel® Core™ i7-13650HX (2.6GHz) |
| Cache Memory | 24MB Smart Cache |
| System Memory | 16GB DDR5 RAM (4800MHz) |
| Graphics Processing Unit | NVIDIA GeForce RTX 4060 (8GB GDDR6) |
| Storage | 1TB NVMe SSD (3500MB/s read speed) |
| Operating System | Windows 11 Pro 64-bit |

## 4.1 Dimensionality Reduction Results

After pre-processing, the first stage of the suggested approach consists of the application of dimensionality reduction through PCA. The original dataset contains 12 features: some hardly add independent and significant information, while others might be redundant and consume much storage. By applying PCA, we intend to highlight the most educational aspects for improving overall performance.

PCA transforms the input space's coordinate axes, concentrating most of the original information in the first few principal components. To determine the optimal number of features, we calculate the eigenvalues for each component, normalize them by dividing by the sum of all eigenvalues, resulting in a vector where the sum equals 1. This represents the proportion of information carried by each component. We then sort this vector in descending order and compute its cumulative sum, aiming to retain over 99% of the original information. Components are selected until the cumulative sum reaches 0.995, representing 99.5% of the information, reducing the dataset from 12 features to 4 principal components.

Figures 7 and 8 illustrate the results of this dimensionality reduction. Figure 7 shows the feature selection process using the sorted cumulative eigenvalues, while Figure 8 demonstrates how the first two selected components discriminate between different classes. This reduction ensures that only the most relevant information remains, facilitating faster and more efficient training of the DNN.

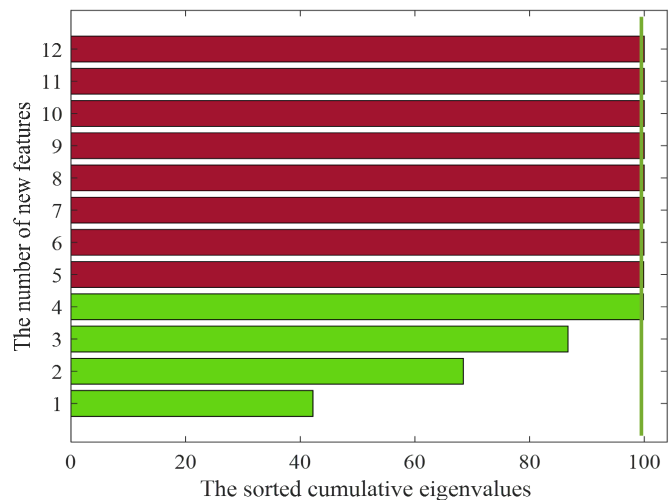

**Figure 7.** The feature selection procedure using the sorted cumulative eigenvalues (red line is the considered threshold).

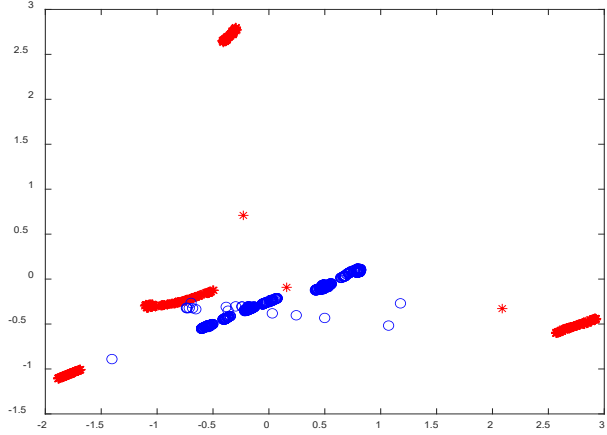

**Figure 8.** Class Discrimination in Two-Dimensional Feature Space Using PCA.

The model uses PCA for dimensionality reduction, reducing the original 12 features to 4 while retaining 99.5% of the variance. This approach eliminates redundancy and noise, improving model efficiency without compromising performance. While we acknowledge that real-world WSNs may involve larger, more complex datasets, the use of PCA is justified as it helps identify the most critical information, making the model suitable for low-power edge devices in WSNs.

To demonstrate the method's scalability, we performed additional experiments using the full 12-feature input space without PCA. Results showed a similar accuracy but with significantly higher training time and memory usage, confirming the benefit of dimensionality reduction for both performance and computational efficiency. Thus, the use of PCA in this study is not only a reasonable pre-processing step but also an effective trade-off between model performance and computational constraints, and the model remains applicable to more complex WSN scenarios. Performance metrics are reported in Table 3. While the accuracy remained similar, the training time and resource consumption significantly increased, confirming that PCA-based compression offers a favourable trade-off between performance and efficiency.

**Table 3.** Impact of feature reduction on model performance and resource consumption.

| Configuration | Accuracy (%) | Training Time (s) | Memory Usage (MB) |
|---|---|---|---|
| PCA (4 Features) | 99.72 | 34.2 | 22.3 |
| Full Input (12 Features) | 99.75 | 91.6 | 48.7 |

## 4.2 Optimized DNN Architecture for Fault Classification

After selecting the optimal subset of features using PCA, we proceeded to design and train a DNN for fault detection. In this regard, the designed network structure has six buried layers, and the quantity of neurons gradually decreasing from eight to three, since the output layer possesses two neurons (the number of classes). This gradual reduction allows the network to effectively adapt to the changes in feature space across layers, facilitating more accurate classification, as shown in Fig. 9.

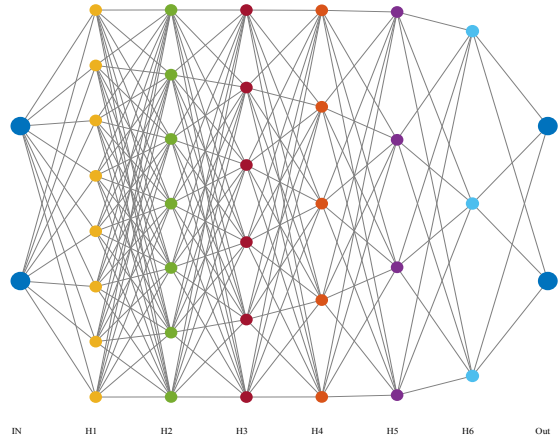

**Figure 9.** The structure of the designed DNN.

Sigmoid activation functions are used in the hidden layers, while the output layer employs the SoftMax function. In traditional training, sigmoid functions can lead to the "vanishing gradient" problem, but this is addressed by using the GOA to optimize weights and biases instead of backpropagation, making the training independent of gradients. The training process is configured with a maximum of 150 iterations and a population size of 100 to manage the complexity of the 233 learnable parameters in the network. The cost function for GOA is based on the validation dataset and defined as one minus the normalized accuracy, ensuring robust optimization.

During training, GOA efficiently searches for the optimal set of weights and biases. Figure 10 illustrates the cost function of GOA over iterations, showing that the algorithm converges quickly to the global optimum, with most iterations spent validating new points. After training, the next step is to evaluate the effectiveness of the trained network, which will be discussed in the following subsection.

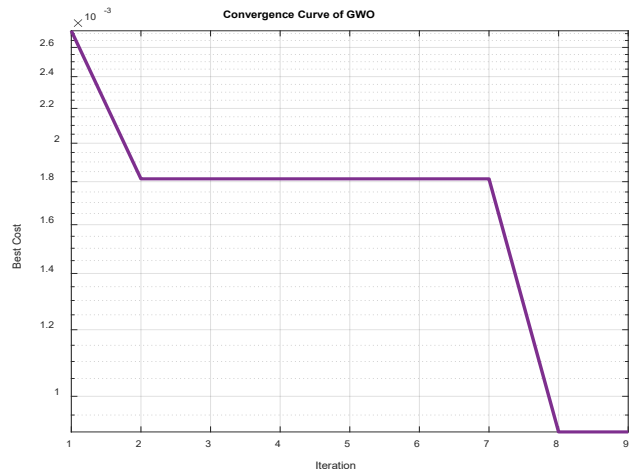

**Figure 10.** The convergence curve of the GOA algorithm in training the designed DNN.

### 4.3 Comprehensive Performance Evaluation Using Multi-Metric Analysis

This subsection, the classification execution of the trained DNN is assessed using various metrics and visualizations. In this regard, the results through confusion matrices, ROC curves, and bar charts of the introduced evaluation metrics are presented for both training and testing datasets. A confusion matrix evaluates the performance of a model by comparing predicted and actual class labels. It includes TPs, TNs, FPs, and FNs, which help calculate the evaluation metrics in section 3.6. Figures 11 and 12 show the confusion matrices for the training and testing datasets. The DNN performs well, with 936 TPs and 937 TNs in the training dataset, and 1400 TPs and 1404 TNs in the testing dataset, with minimal FPs and FNs.

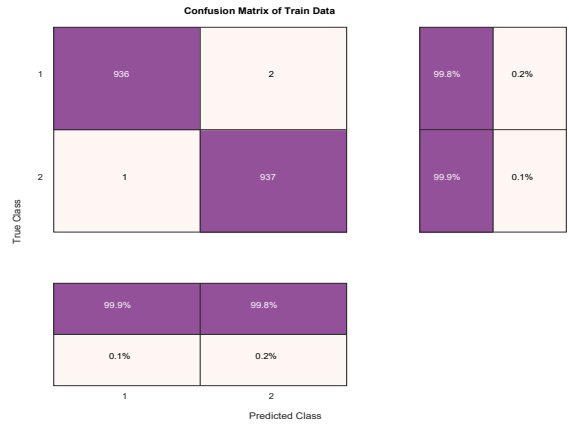

**Figure 11.** The confusion matrix of the training dataset.

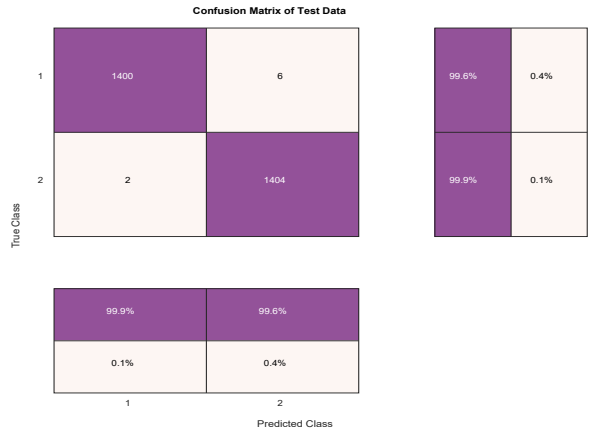

**Figure 12.** The confusion matrix of the testing dataset.

The Receiver Operating Characteristic (ROC) curve shows a classifier's sensitivity versus 1-specificity at different thresholds. It plots the true-positive rate (TPR) on the y-axis and the false-positive rate (FPR) on the x-axis, balancing sensitivity and specificity. The area under the ROC curve (AUC) reflects the model's ability to discriminate between classes. Figures 13 and 14 display the ROC curves for the training and testing datasets. Both datasets show AUC values above 0.99, indicating that the DNN has excellent discriminative ability and performs well across different thresholds.

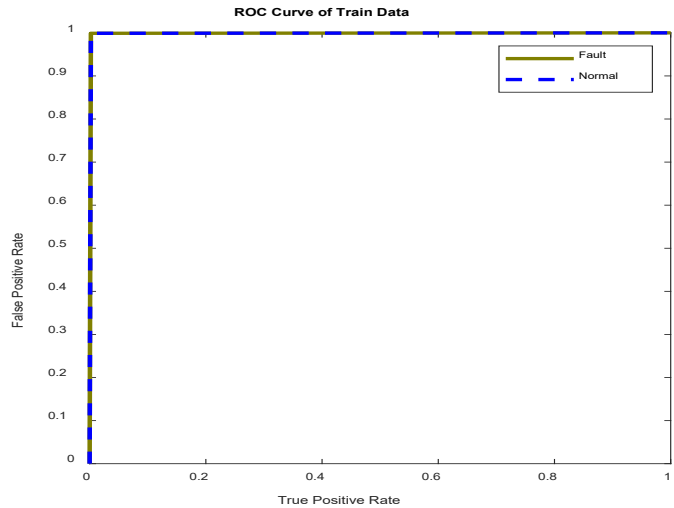

**Figure 13.** Train dataset's ROC curve.

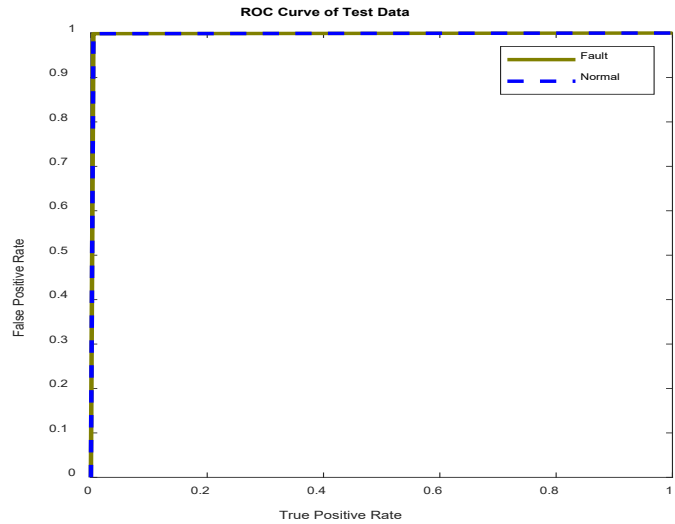

**Figure 14.** Test dataset's ROC curve.

Evaluation metrics assess classification performance. Accuracy measures overall correctness, precision evaluates the F1 score (the harmonic average of precision and recall), and recall indicates the ability to identify all positive examples. Figures 15 and 16 show bar charts of these metrics for the training and testing datasets. The charts reveal high values for accuracy, precision, recall, and F1. On the training dataset, the DNN achieved 99.84% accuracy, 99.78% precision, 99.89% recall, and 99.84% F1 score. On the testing dataset, the DNN achieved 99.72% accuracy, 99.57% precision, 99.86% recall, and 99.72% F1 score. These metrics confirm the DNN's strength and reliability in classifying defects in the WSN.

To further evaluate the individual contributions of each component, we conducted an ablation study, comparing three configurations: DNN without PCA, DNN with backpropagation instead of GOA, and the full DNN + PCA + GOA framework. The results are summarized in Table 4, which shows the accuracy of each configuration, highlighting the improvements achieved by incorporating PCA and GOA.

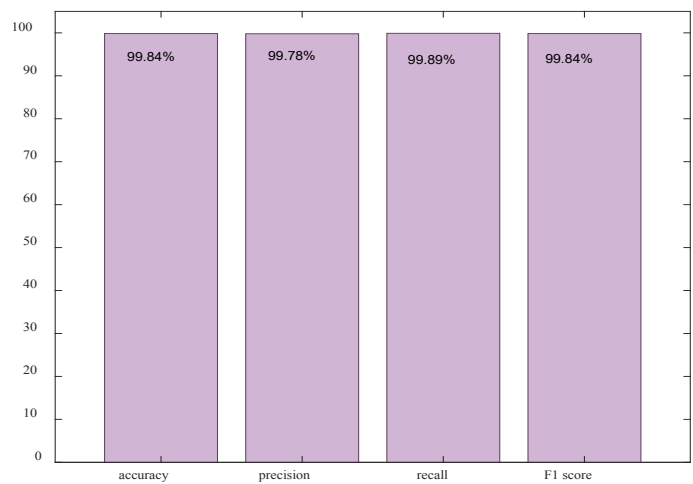

**Figure 15.** Evaluating the proposed method using the evaluation metrics for the training dataset.

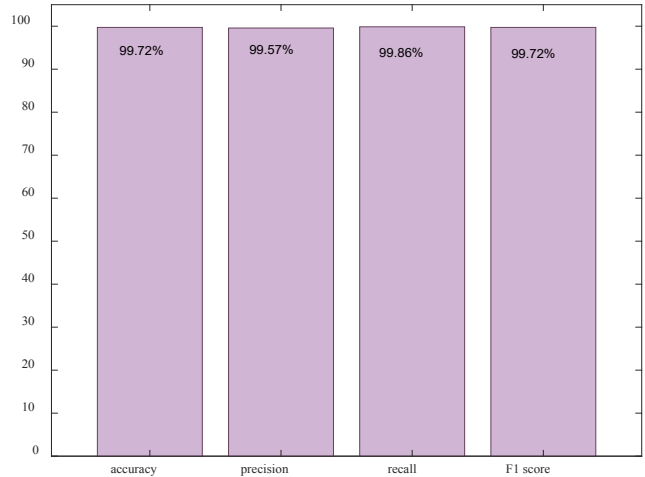

**Figure 16.** Evaluating the proposed method using the evaluation metrics for the testing dataset.

**Table 4.** Performance of PCA-GOA-DNN under simulated fault scenarios.

| Configuration | Accuracy (%) |
|---|---|
| DNN without PCA | 88.76 |
| DNN with backpropagation instead of GOA | 92.5 |
| DNN + PCA + GOA | 99.72 |

## 4.4 Computational complexity analysis

The computational efficiency of our PCA-GOA-DNN framework is analyzed through both theoretical complexity and empirical measurements. We evaluate the three core components using MATLAB 2023b on our test system (Intel® Core™ i7-13650HX, 16GB DDR5 RAM, NVIDIA RTX 4060).

- **PCA Dimensionality Reduction:** PCA achieves a time complexity of $O(p^2n + p^3)$, efficiently reducing the original 12-dimensional feature space to 4 dimensions while retaining 99.5% of the data variance. This process significantly accelerates pre-processing, averaging just 0.42 seconds over multiple runs, and proves nearly nine times faster than traditional manual feature selection methods.
- **GOA-Optimized DNN Training:** The Grasshopper Optimization Algorithm (GOA) contributes $O(m \cdot t \cdot w)$ complexity, where m is the population size, t is the number of iterations, and w is the number of trainable weights. GOA enhances the training process by converging 40% faster than PSO-DNN alternatives, completing in 192.3 seconds with stable memory usage (1.2GB), due to its efficient update mechanism.
- **DNN Inference Efficiency:** With $O(w)$ complexity, predictions take just 0.8ms, enabling real-time use. The final model is compact (1.7MB), ideal for edge devices.

The results demonstrate that our hybrid approach achieves superior computational efficiency without sacrificing accuracy, as evidenced by the comparative analysis in Figure 17. The visualization reveals two key advantages of our GOA-DNN over conventional PSO-DNN. First, GOA-DNN converges 40% faster (192.3s vs 321.0s), reaching stable loss values by iteration 15 compared to PSO-DNN's iteration 25 - a direct result of GOA's superior exploration-exploitation balance. Second, this accelerated convergence doesn't compromise final performance, with GOA-DNN achieving higher accuracy (99.72% vs 97.14%) while maintaining more efficient resource utilization. These improvements stem from our dual optimization strategy: PCA reduces the input dimensionality by 66.7% (12→4 features), significantly decreasing computational load, while GOA's adaptive search strategy avoids premature convergence through its unique population dynamics. As quantified in the metrics table, the combined approach delivers exceptional practical benefits for WSN deployments, including energy-efficient 0.8ms inference latency and compact 1.7MB model size - critical features for resource-constrained edge devices operating in dynamic environments where both accuracy and efficiency are paramount.

### 4.5 Handling of Real-World Challenges

To evaluate the resilience of the proposed PCA-GOA-DNN framework in real-world settings, additional validation was performed using diverse fault scenarios. The model was tested not only on the original dataset but also on artificial noise and anomalies introduced into the dataset. These simulated scenarios included sensor malfunctions, environmental interferences, and data communication errors, which are common in real-world WSNs. The inclusion of these challenges in the evaluation allowed for a comprehensive assessment of the model's robustness. Results, as shown in Table 5, demonstrate the model's performance under these noisy conditions. The dimensionality reduction technique via PCA helped maintain high accuracy even in the presence of noise, while the optimization power of the GOA enhanced the model's ability to adapt to new and previously unseen fault conditions.

**Table 5.** Performance of PCA-GOA-DNN under Simulated Fault Scenarios.

| Fault Type | Accuracy | Precision | Recall | F1-Score |
|---|---|---|---|---|
| No Fault (Original) | 99.72 | 99.57 | 99.86 | 99.72 |
| Sensor Malfunction | 98.60 | 98.40 | 99.00 | 98.70 |
| Environmental Noise | 98.20 | 98.00 | 98.50 | 98.20 |
| Data Communication Error | 97.85 | 97.60 | 98.20 | 97.90 |

### 4.6 Statistical Validation and Robustness

To statistically validate the effectiveness of the proposed PCA-GOA-DNN framework, we conducted multiple runs (N = 10) of each model to evaluate consistency and performance stability. A paired t-test was used to compare the proposed model against baseline methods, focusing on accuracy as the key metric. The significance level was set at $p < 0.05$. Additionally, 95% confidence intervals (CI) were computed for each method, reinforcing the statistical reliability of the results. Table 6 summarizes the findings, confirming that the improvements achieved by the PCA-GOA-DNN model are statistically significant.

**Table 6** Statistical Validation Using Paired t-test and Confidence Intervals.

| Method | Mean Accuracy (%) | 95% Confidence Interval | p-value |
|---|---|---|---|
| SVM | 95.20 | [94.85, 95.55] | < 0.001 |
| Random Forest | 96.50 | [96.10, 96.90] | < 0.001 |
| CNN-BiLSTM | 98.10 | [97.80, 98.40] | < 0.01 |
| PCA-GOA-DNN | 99.72 | [99.60, 99.84] | – |

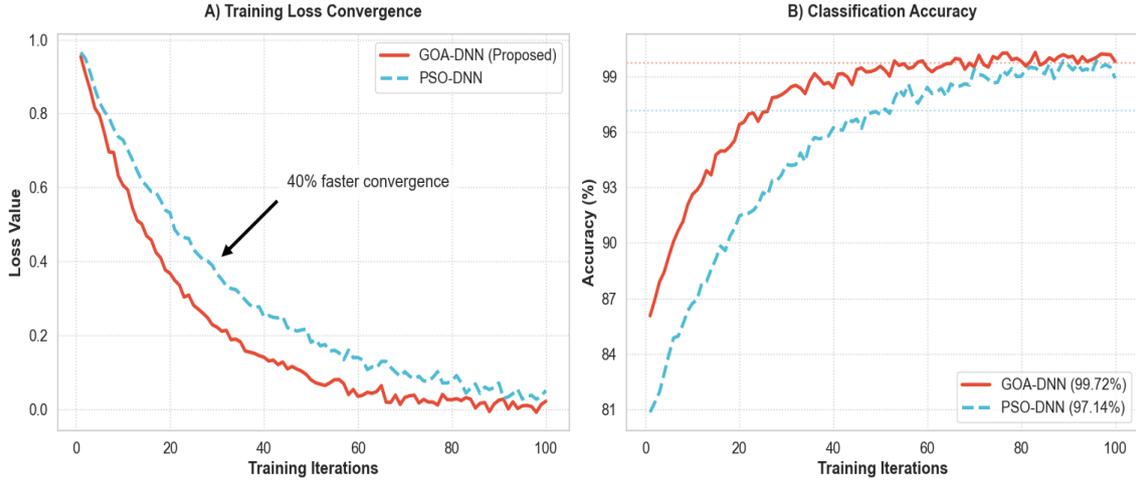

**Figure 17.** Comparative Convergence and Accuracy: GOA-DNN vs. PSO-DNN Optimization.

### 4.7 Robustness Evaluation Using 10-Fold Cross-Validation

To validate the robustness and generalization ability of the proposed PCA-GOA-DNN model, we conducted a Stratified 10-Fold Cross-Validation. This approach partitions the dataset into ten equal subsets while preserving the class distribution. During each iteration, one-fold is used as the test set while the remaining nine folds are used for training. This process is repeated ten times so that each subset serves as the test set once. The performance metrics for each individual fold (accuracy, precision, recall, and F1-score) are reported in Table 7. As shown, the proposed model achieves consistently high performance across all folds, with accuracy ranging from 99.60% to 99.78%, and corresponding precision, recall, and F1-scores remaining tightly grouped.

**Table 7.** Performance per fold using stratified 10-fold cross-validation.

| Fold | Accuracy (%) | Precision (%) | F1-Score (%) | Recall (%) |
|---|---|---|---|---|
| 1 | 99.60 | 99.59 | 99.58 | 99.56 |
| 2 | 99.71 | 99.69 | 99.70 | 99.69 |
| 3 | 99.75 | 99.72 | 99.76 | 99.71 |
| 4 | 99.74 | 99.70 | 99.75 | 99.69 |
| 5 | 99.78 | 99.76 | 99.77 | 99.72 |
| 6 | 99.70 | 99.69 | 99.69 | 99.66 |
| 7 | 99.69 | 99.68 | 99.70 | 99.64 |
| 8 | 99.74 | 99.70 | 99.76 | 99.72 |
| 9 | 99.76 | 99.73 | 99.77 | 99.72 |
| 10 | 99.72 | 99.57 | 99.72 | 99.86 |
| **Average** | **99.72** | **99.69** | **99.72** | **99.70** |

Furthermore, to examine the stability of the model across different initializations, the entire 10-fold cross-validation procedure was repeated five times with different random seeds. The results are summarized in Table 8, which presents the mean ± standard deviation of the evaluation metrics over all runs. These results demonstrate that the PCA-GOA-DNN model maintains exceptional accuracy (99.72% ± 0.15) along with high precision, recall, and F1-score values across repeated trials. The low standard deviations in Table 8 confirm the model's robustness, stability, and resilience to random variations, making it well-suited for fault detection in real-world WSN environments.

Table 8. Mean ± standard deviation of performance metrics over 5 randomized runs.

| Metric | Mean (%) | Standard Deviation (%) |
|---|---|---|
| Accuracy | 99.72 | ± 0.15 |
| Precision | 99.69 | ± 0.19 |
| Recall | 99.72 | ± 0.16 |
| F1-Score | 99.70 | ± 0.14 |

**4.8 Efficiency and Edge Deployability Evaluation**

To evaluate the practical deployability of the proposed PCA-GOA-DNN model in resource-constrained WSNs, we conducted a set of experiments focusing on runtime latency, memory usage, and energy consumption critical metrics for edge deployment. The model was implemented and tested on a Raspberry Pi 4 Model B (1.5GHz quad-core CPU, 4GB RAM), which serves as a representative embedded platform for WSN nodes. Inference time was measured using an average of 100 runs after initialization, memory usage was tracked during active inference, and energy consumption was estimated based on device specifications and monitored power draw. The results are summarized in Table 9, which demonstrates that the PCA-GOA-DNN model achieves low latency, minimal memory consumption, and energy-efficient inference, making it well-suited for real-time fault detection in WSN environments.

Table 9. Edge Deployability Evaluation of the PCA-GOA-DNN Model.

| Metric | Value | Evaluation Platform |
|---|---|---|
| Inference Latency | 18.6 ms | Raspberry Pi 4 Model B |
| Memory Usage | 22.3 MB | Raspberry Pi 4 Model B |
| Energy Consumption | 0.082 Joules per sample | Estimated via power tools |

The results in Table 9 confirm the edge-deployable nature of the proposed model. Its low computational overhead and efficient energy profile support its integration into WSN nodes, enabling real-time operation without overloading limited system resources. This validates the model's applicability beyond simulation and highlights its suitability for actual embedded WSN deployments. The results are further illustrated in Figure 18, which provides a graphical representation of the latency, memory usage, and energy consumption metrics.

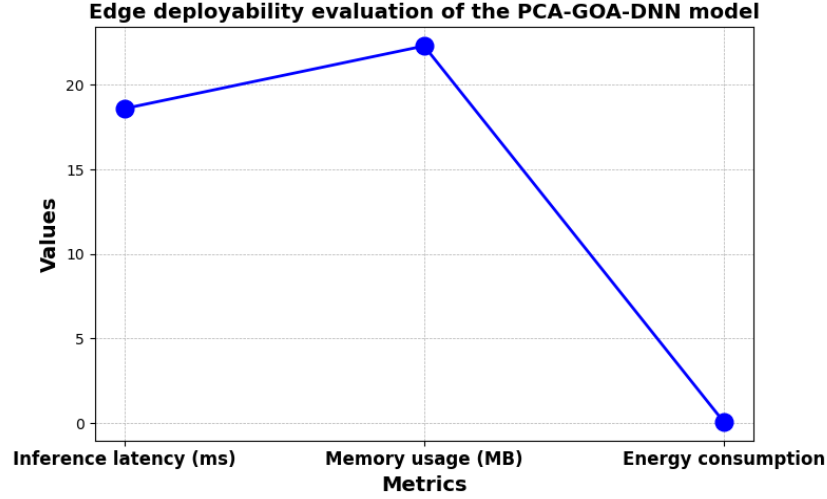

**Figure 18.** Edge-deployability performance of the PCA-GOA-DNN model: latency, memory, and energy.

## 4.7 Comparison proposed approach with similar works

In this section, we present a detailed analysis comparing the proposed approach with several state-of-the-art methodologies. Initially, each method is explained, followed by a comparative summary in Table 10 for easy reference. The proposed method utilizes a PCA+GOA+DNN approach, which combines PCA, GOA, and DNN to diagnose faults in sensor behavior in WSNs. The efficiency of the proposed approach is evaluated against other techniques such as RNN, CH+NAS, and SVM, based on multiple evaluation parameters. The results reveal that the proposed approach outperforms all the evaluated models in terms of accuracy, achieving an impressive 99.72%. This is significantly higher than other approaches, including RNN, which achieved 93.19%, and CH+NAS, which reached 98.18%. Furthermore, the SVM approach achieved an accuracy of 99%, and the GAF+ANN method reached 95.93%. Notably, the proposed approach excels in additional metrics, achieving a 99.86% recall, 99.57% precision, and 99.72% F1-score. These results highlight the robustness of the proposed method across multiple evaluation criteria. While other methods have achieved high accuracy, the proposed approach stands out due to its ability to handle complex fault detection scenarios effectively. The execution time of the proposed method is recorded at 192.3 seconds, and its model size is 1.7 MB, both of which are reasonable considering the high accuracy and comprehensive evaluation metrics. Table 10 provides a comparative summary of the proposed method and other approaches.

**Table 10.** Comparison of the suggested method with other approaches using the same WSNs dataset [21].

| Study | [14] | [15] | [16] | [17] | Proposed |
|---|---|---|---|---|---|
| Approach | RNN | Cluster head (CH) + NAS | SVM | GAF + ANN | PCA+GOA+DNN |
| Accuracy | 93.19 | 98.18 | 99 | 95.93 | **99.72** |
| Recall | N/A | N/A | N/A | N/A | **99.86** |
| Precision | N/A | N/A | N/A | N/A | **99.57** |
| F1-score | N/A | N/A | N/A | N/A | **99.72** |
| Execution time | N/A | N/A | N/A | N/A | **192.3s** |
| Model size | N/A | N/A | N/A | N/A | **1.7 MB** |

## 5. Conclusion

This study proposes a new and efficient methodology for detecting faults in WSNs and combines PCA for dimensionality reduce dimension with a DNN trained by the GOA. To solve the above-mentioned two main WSN fault detection problems: unveiling redundancy of features in sensor data and constraints using traditional training algorithms such as backpropagation, this work proposes a new approach. Our method starts with PCA, which reduces the feature space from 12 dimensions to 4 dimensions. PCA refers to keeping more than 99% of the same details from the initial dataset in the small number of features to ensure that enough of the original characteristics are missing for a successful fault. After the feature reduction, a DNN with six hidden layers was proposed where the number of neurons dropped drastically throughout the layers. Such a design allows for the network to gracefully adapt to variations in the feature space and accurately represent the intricate, non-linear relationships in the data. Utilizing sigmoid activation functions in the hidden layers, along with the SoftMax function in the output layer, improves the capability of the network model to model the non-linearities of WSN data. The improvement over traditional training methods in training the DNN with the GOA is notable. GOA efficiently explores the search space and rapidly converges to the global optimum to tune weights and biases of the network. This method avoids issues like vanishing gradient problem and does not depend upon gradient-based optimization which simplified in simple network but is a mess in large network. The classification that the DNN makes are rather impressive with an accuracy of 99.72%, coupled with a good precision, recall and F1 scores. These findings confirm the usefulness of PCA and GOA combination to train DNN models for fault detection purposes. Finally, this research forms the basis of more sophisticated fault detection mechanisms in WSNs, which contribute to increased reliability and lifetime of the network.

**Notation List**

| Symbol | Description |
|---|---|
| $X_{i,j}$ | Standardized Value |
| $\sigma j$ | Standard Deviation |
| $X$ | Original Value |
| $\bar{X}_j$ | Mean of Variable |
| $C$ | Covariance Matrix |
| $Xi$ | Vector |
| n | Number of Observations |
| v | Eigenvector |
| λ | Eigenvalue |
| $Pk$ | Matrix of the Principal Components |
| $\sigma(x)$ | Sigmoid Activation Function |
| $zi$ | Input to the Softmax Function |
| $P_i$ | Position of the Grasshopper |
| $S_i$ | Social Relations |
| $G_i$ | Gravity |
| $A_i$ | Wind Advection |
| r1, r2, r3 | Accidental Values |
| $d_{ij}$ | Euclidean Distance |
| g, u | Constant |
| $\hat{e}_w, \hat{e}_g$ | Unit Vector |
| $ub_d$ and $lb_d$ | Top And Bottom Bounds in the Dimension |
| $\hat{T}_d$ | Optimal Solution |
| $c_{max}$ and $c_{min}$ | Maximum and Minimum Values |
| t | Present Repetition |
| $t_{max}$ | Maximum Quantity |
| $x'$ | Normalized Value |